# 考虑可执行度的地铁服务质量提升策略研究

陈维亚[1†]　李嘉佳[1]　康梓轩[1]

（1．中南大学 交通运输工程学院，轨道交通大数据湖南省重点实验室，湖南 长沙 410075）

**摘　要**：基于地铁运营服务质量评价结果制定合理的策略提升服务质量，是地铁运营管理的重要工作。本文融合考量一定时期内服务质量指标的得分、权重和改进可执行度，将决策树引入重要性战略矩阵，构建了用于决策服务质量指标改进优先级和改进幅度的决策树-重要性战略矩阵模型（DT-IPA）。该模型通过构建决策树生成分类规则，运用层次分析法计算指标改进可执行度，以此为基础优化由IPA确定的指标初始改进优先级并量化被调整指标的改进幅度，实现调整后的总体服务质量达到预期目标。以长沙地铁为例，验证了DT-IPA模型的有效性。该方法可用作地铁运营单位提升服务质量的决策工具。

**关键词**：地铁；服务质量；提升策略；可执行度；决策树

**中图分类号**：U231+.92　　**文章编号**：1000-565X

根据我国交通部2019年公布施行的《城市轨道交通服务质量评价规范》[1]，城市轨道交通运营单位应当按照有关标准为乘客提供安全、可靠、便捷、高效、经济的服务，保证服务质量，采用先进技术提升服务品质；城市轨道交通运营主管部门应当通过乘客满意度调查等多种形式，定期对运营单位服务质量进行监督和考评。因此，运营单位根据运营服务质量评价结果制定合理的策略提升服务质量，是地铁运营管理的重要工作。

在根据指标评价结果制定服务质量提升策略时，由于受到技术条件、资源等限制，地铁运营单位往往难以在一段时期内同时提升所有指标，因此需要决策评价指标的改进优先级和改进幅度。重要性战略矩阵（Importance-Performance Analysis，IPA）是目前常用的决策方法[2]，它通过同时考虑指标的服务质量得分和权重，把指标划分为优先改进区、保持优势区、次要改进区和锦上添花区四个类别，对应四种改进优先级[3]。但IPA的决策过程如未考虑指标改进的可执行度，决策结果难以实施和保障效果。例如，张慧慧[4]和梅家骏[5]通过IPA发现车厢拥挤状况是北京地铁和广州地铁优先改进区的指标。但是，北京地铁和广州地铁的客流量巨大，列车发车频率已基本达到极限，因此在一定时期内难以通过调整发车频率等运营管理手段增大运输能力和降低拥挤度。

为了弥补当前的IPA决策不考虑指标改进可执行度的缺陷，本文在IPA决策模型中引入决策树（Decision Tree, DT），构建DT-IPA决策模型，实现融合考量指标的得分、权重和可执行度，决策出合理的服务质量提升策略，并以长沙地铁为例开展实证研究。

## 1 决策模型构建

### 1.1 模型框架

本文的决策问题是：根据地铁服务质量评价指标体系及评价得分，融合考量指标的权重及改进可执行度，确定指标的改进优先级和改进幅度，以实现预期的质量提升目标。如图1所示，构建DT-IPA模型的输入数据包括：指标体系和指标得分。决策过程分五步：第一步，利用指标体系和指标得分确定最佳决策树模型；第二步，基于最佳决策树计算指标权重；第三步，利用指标的权重和得分均值构建IPA，第四步，通过层次分析法（Analytic Hierarchy Process，AHP）量化改进指标的可执行度；第五步，针对落在IPA优先改进区且无改进可执行度的指标，利用决策树中包含相关指标的分类规则，调整分类规则内指标的改进优先级并确定改进幅度，使调整后总体服务质量得分能达到较高等级。最后，DT-IPA模型将输出指标的改进优先级和改进幅度。以下详细阐述DT-IPA模型的决策步骤。

收稿日期：
基金项目：湖南省自然科学基金资助项目(2018JJ2537)
Foundation items: Supported by the Natural Science Foundation of Hunan Province (2018JJ2537)
作者简介：陈维亚(1981-)，男，湖南桃江人，副教授，博士，从事交通运输规划与管理研究；E-mail：wychen@csu.edu.cn

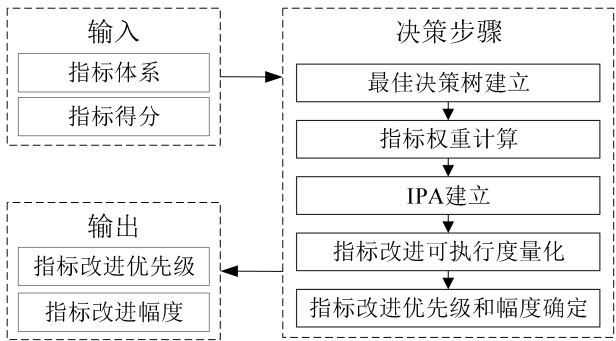

图 1 DT-IPA 模型框架

Fig.1 Framework of DT-IPA model

## 1.2 关键步骤

### 1.2.1 最佳决策树建立

决策树为由一系列的节点分裂而成的树状分类或回归模型。按节点分裂方法分类，建立决策树的算法分为 ID3、C4.5 和 CART 等。由于 CART 的生成速度和分类准确率优于 ID3 和 C4.5[6]，本文选用 CART 算法。另外，因为地铁乘客服务质量的测评数据是离散的，所以运用 CART 算法建立的决策树为分类树，输入数据 $\Omega$ 是各指标得分和总体服务质量得分，如式所示。获得输入数据后，通过图 2 所示四个步骤建立最佳决策树。

$$\Omega = \{D_1, D_2, ..., D_M\} = \begin{bmatrix} x_{11} & x_{12} & ... & x_{1M} \\ x_{21} & x_{22} & ... & x_{2M} \\ ... & ... & x_{nm} & ... \\ x_{N1} & x_{N2} & ... & x_{NM} \end{bmatrix} \quad (1)$$

式中，行变量为被访者，列变量为服务质量评价指标；$x_{nm}$ 表示被访者 $n$ 对指标 $m$ 的打分，$D_m$ 为指标 $m$ 得分集合，$D_m = \{x_{1m}, x_{2m}, ..., x_{Nm}\}$ $n = 1, 2, ..., N$，$N \in N^*$，$m = 1, 2, ..., M$，$M \in N^*$。

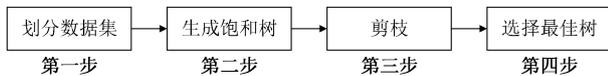

图 2 最佳决策树生成基本步骤

Fig.2 Basic steps for building an optimal Decision-Tree

第一步，运用 $k$ 折交叉验证划分训练集数据和测试集数据。其中，$k$-1 组为训练集数据，用于生成饱和树；剩余的 1 组为测试集数据，用于测试剪枝树的分类准确度。数据集将进行 $k$ 次划分，重复以上操作。大量的实验证明当等于 10 时，分类准确度最好[7]。因此，本文令 $k$ 等于 10。

第二步，通过 CART 算法，使用训练集数据生成饱和树。CART 算法定义 Gini 系数为节点数据类别不一致的概率，如式所示。

$$Gini(D) = 1 - \sum_{k=1}^{K} (\frac{|C_k|}{|D|})^2 \quad (2)$$

式中，$D$ 表示节点的数据集，是按某服务质量评价指标的某个得分划分的数据集，如"站内指引标志"得分大于 2 的数据集，$|D|$ 表示数据集的数据量；$k$ 表示得分；$|C_k|$ 表示数据集 $D$ 含有总体服务质量得分为 $k$ 的数据量。

基于 CART 算法生成的决策树为二叉树，其分裂思想是最小化子节点的 Gini 系数，即选择合适的分裂器使子节点的数据尽量属于同一个类别[6]。每一次分裂节点减小的 Gini 系数如式所示。

$$Gini(D, A) = Gini(A) - [\frac{|D_1|}{|D|}Gini(D_1) + \frac{|D_2|}{|D|}Gini(D_2)] \quad (3)$$

式中，$A$ 表示父节点的数据集；$D$ 为子节点的数据集，$|D|$ 为子节点数据集含有的数据量。

通过递归，决策树的 Gini 系数会随着树的生长逐渐减小。当 Gini 系数无法减小或叶子节点所含的数据量达到设置的最小值时，算法停止，获得饱和树[8]，记为 $T_0$。

第三步，利用代价复杂度算法剪枝，目的是避免决策树出现过拟合现象。代价复杂度剪枝的基本思想是：用一个叶子节点代替饱和树内某个子树，生成一棵剪枝树。对这颗树执行相同的操作，递归，生成一系列的剪枝树 $T_0, T_1, T_2, ..., T_n$。最后一棵生成的剪枝树 $T_n$ 仅有一个根节点。

对树 $T_i$ 的每个非叶子节点 $t$ 都执行式的运算，找到 $g(t) = \alpha$ 对应的节点 $t$，令其左、右子节点为 null，生成 $T_{i+1}$。当多个非叶子节点的 $g(t)$ 同时达到最小时，取 $|T_t|$ 最大的节点 $t$ 进行剪枝。

$$g(t) = \frac{C(t) - C(T_t)}{|T_t| - 1} \quad (4)$$
$$\alpha = \min(\alpha, g(t))$$

式中，$|T_t|$ 和 $C(T_t)$ 分别是以节点 $t$ 为根节点的树的叶子节点数量和预测误差，$C(t)$ 为剪枝后节点 $t$ 的预测误差。

第四步，利用测试集数据从一系列剪枝后的决

策树选出最佳决策树。定义所有叶子节点的 $Gini$ 系数之和为决策树的分类误差率[9]，如式所示。

$$\text{分类误差率} = \sum_{l=1}^{L} \frac{|D_l|}{|D|} Gini(D_l) = \sum_{l=1}^{L} \frac{|D_l|}{|D|}[1 - \sum_{k=1}^{K}(\frac{|C_k|}{|D_l|})^2] \tag{5}$$

式中，$l$ 表示叶子节点的编号，$L$ 表示叶子节点的数量，其余符号的含义与式一致。

运用 $k$ 折交叉验证划分的每个对折的测试集数据，计算 $T_0, T_1, T_2, ..., T_n$ 中每棵决策树的分类误差率。每棵决策树因此可获 $k$ 个分类误差率，把 $k$ 个分类误差率的平均值作为该树的分类误差率。一个简单的原则是选择交叉验证分类误差率最低的树。若简单的树和复杂的树的交叉验证分类误差大致相同，应优先选择简单的树，有利于使决策树避免过拟合。决策树的规模一般使用叶子节点数量表示[8]。叶子节点数量越多，决策树的规模越大。通过绘制剪枝树 $T_0, T_1, T_2, ..., T_n$ 的分类误差率-叶子节点数量的图像，距离最小分类误差一个标准差的最小规模的剪枝树即为最佳树。

图 3 展现了基于输入数据式，利用 CART 算法构造的决策树的一般结构。它包含根节点、子节点、叶子节点和分裂器，节点用长方形表示，分裂器用菱形表示。根节点是所有评价指标的得分数据；总体服务质量指标以外的评价指标充当分裂器，按指标得分分裂；叶子节点代表总体服务质量的分类情况，其余节点表示其它评价指标的分类情况[8]。

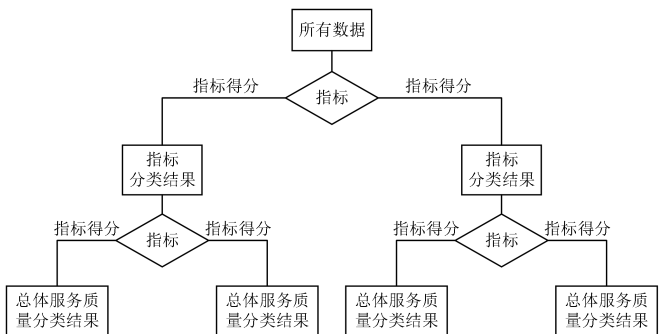

图 3 基于地铁服务质量评价指标得分数据建立的决策树一般结构

Fig.3 General structure of decision tree based on score data of metro service quality attributes

### 1.2.2 指标权重计算

根据 Kashani 和 Mohaymany [9]，指标权重是决策树的节点按该指标分裂后，节点 $Gini$ 系数减少的加权平均值（权重是分裂器所含样本量与所有数据样本量之比）。指标 $i$ 的权重 $w(i)$ 的计算式如式所示。

$$w(i) = \sum_{t=1}^{T} \frac{n_t}{N} * \{Gini(t) - [\frac{|D_1|}{|D|}Gini(D_1) + \frac{|D_2|}{|D|}Gini(D_2)]\} \tag{6}$$

式中，$t$ 表示以指标 $i$ 为分裂器的节点，节点数量可能大于 1；$T$ 表示决策树中所有节点的数量；$n_t$ 表示以指标 $i$ 为分裂器的节点的样本量；$N$ 表示根节点的样本量；$D_1$ 和 $D_2$ 表示节点 $t$ 以指标 $i$ 为分裂器产生的子节点。

### 1.2.3 IPA 建立

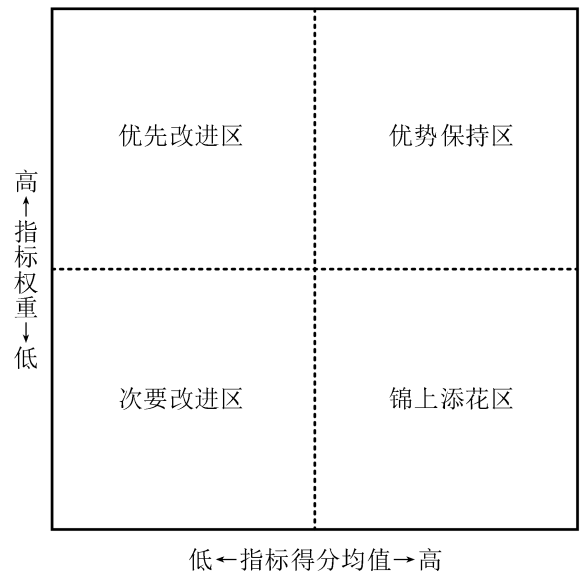

图 4 重要性战略矩阵（IPA）

Fig.4 Importance-Performance analysis, IPA

以指标得分和权重为坐标轴，所有指标得分的均值 $\bar{x}$ 和权重的均值 $\bar{w}$ 为原点，建立如图 4 所示的 IPA。基于所求的指标得分均值和权重，确定各个指标的初始改进优先级。

左上角象限为"优先改进区"，该区域的指标权重大于 $\bar{w}$，得分均值低于 $\bar{x}$，具有最高改进优先级，记为第一优先级。右上角象限为"优势保持区"，该区域的指标得分均值和权重分别大于 $\bar{x}$ 和 $\bar{w}$，需要继续保持优势，具有第二高改进优先级，记为第二优先级。左下角象限为"次要改进区"，该区域

的指标得分均值和权重分别低于 $\bar{x}$ 和 $\bar{w}$，具有第三高改进优先级，记为第三优先级。最后，右下角象限为"锦上添花区"，该区域的指标得分均值大于 $\bar{x}$，而权重低于 $\bar{w}$，具有最低改进优先级，记为第四优先级。

#### 1.2.4 指标改进可执行度量化

指标改进可执行度是指在一定时期内考虑技术条件、资源等条件下，服务质量指标能够被改进的可行性及程度。本文利用层次分析法量化地铁服务质量指标改进可执行度。首先请地铁运营管理专家通过 1~9 标度法（表 1）两两比较评价指标的改进可执行度，根据比较结果构造判断矩阵。然后，计算判断矩阵的特征向量并检验一致性。若判断矩阵通过一致性检验，把归一化处理后的特征向量作为评价指标的改进可执行度；否则，对判断矩阵进行调整，直至通过一致性检验。

表 1 评价指标改进可执行度 1~9 标度含义

Table 1 The 9-point scale of the feasibility for attribute's improvement

| 标度值 | 式子 | 含义 |
| --- | --- | --- |
| 1 | $m_i/m_j$ | 两个指标相比，改进可执行度相同 |
| 3 | $m_i/m_j$ | 两个指标相比，$m_i$ 改进可执行度稍大于 $m_j$ |
| 5 | $m_i/m_j$ | 两个指标相比，$m_i$ 改进可执行度明显大于 $m_j$ |
| 7 | $m_i/m_j$ | 两个指标相比，$m_i$ 改进可执行度强烈大于 $m_j$ |
| 9 | $m_i/m_j$ | 两个指标相比，$m_i$ 改进可执行度极其大于 $m_j$ |
| 2, 4, 6, 8 | $m_i/m_j$ | 上述两相邻判断的中间值 |

#### 1.2.5 指标改进优先级和改进幅度确定

被调整指标的改进幅度，使总体服务质量得分能达到较高等级。从最佳决策树的根节点到每个叶子节点的路径为一条分类规则。

式展现了一个分类规则例子，它可以解读为：当指标 1 得分均值大于等于 $k_1$，指标 2 得分均值小于等于 $k_2$，到指标 $i$ 得分均值小于 $k_i$，则总体服务质量有 $a\%$ 的概率得分为 $k$。

如果(指标1($\geq k_1$)，指标2($\leq k_2$)，...，指标$i$($< k_i$))
→ 则（总体感知质量($=k$)），$a\%$ 的Probability
　　　　　　　　　　　　　　　　　　　　　　　　(7)

决策树生成的分类规则并非全部有效，分类规则的有效性一般使用 Support（S）、Population（Po）和 Probability（P）评价[8]。S 表示分类规则出现的概率，Po 表示分类规则中"如果…"部分出现的概率，P 表示符合叶子节点分类情况的数据量与该叶子节点所含的数据量之比。因此，P 等于 S 比 Po 的值，Po 一般由 P 和 S 计算。不同研究人员给定的 S、Po 和 P 的最小阈值不一，本文使用由 De Oña 等[10]提出的较严格的阈值：S 为 0.006，Po 为 0.01，P 为 0.6。

针对改进可执行度不足的指标，从决策树中提取包含相关指标且部分其他指标得分高于现状的分类规则。在分类规则的"如果…"部分，指标可划分为两类：改进可执行度不足和改进可执行度足够的指标。对于改进可执行度足够的指标，根据指标当前得分是否达到分类规则所述要求，又可划分为达标和不达标两类。对于改进可执行度不足且得分已达标的指标，把它们的改进优先级调整为最低。而对于得分未达标的指标，把它们的改进优先级调整为最高，通过对照指标当前得分和分类规则所述得分要求确定改进幅度。不包含在分类规则的指标则不确定改进幅度。

## 2 实证研究设计与实施

### 2.1 问卷设计

本文通过乘客自填式问卷收集数据，问卷包含三个部分。第一部分，设置过滤问题"您是否曾经搭乘长沙地铁"筛选长沙地铁的乘客。若被访者没有乘坐过长沙地铁，调查直接结束。第二部分，收集地铁服务质量评价指标得分。首先通过阅读相关文献[1,11–15]，从安全、舒适和便捷三个维度筛选出 18 个评价指标，分别是：车站可达性、站内指引标志、购票与充值服务、进出站闸机等候、线路图信息、扶手梯与升降梯、站内拥挤度、列车到站信息、候车时长、车厢拥挤度、噪声、照明、温度与通风、卫生、员工服务、生命与财产安全、运营时长和总体服务质量评价。各评价指标得分通过 1 至 5 分的五级数值量表测度，1 分表示非常差的服务质量，5 分表示非常好的服务质量。第三部分，收集被访者个人社会属性、出行特征与本次评价针对的运营服务时段信息，目的是了解样本的代表性和全面性。

### 2.2 调查实施与检验

调查于 2020 年 11 月在长沙轨道交通运营有限公司开展，共收回 130 份问卷。剔除重复提交、存在漏答题项的问卷后，获得有效问卷 107 份，大于所需最小样本量 96（95%置信度，±10%误差）。有效问卷的被访乘客的个人社会属性、出行特征与评

价的运营服务时段覆盖面较全、比例均匀，说明样本的数据的代表性较高。

利用克朗巴哈α系数检验问卷信度，即问卷测度地铁服务质量结果的稳定性[16]。计算结果为 0.964，大于 Devon 等[16]和 Li 等[17]所述的阈值 0.7，说明本次实证研究所获数据具有良好的信度。

# 3 结果分析

## 3.1 最佳决策树

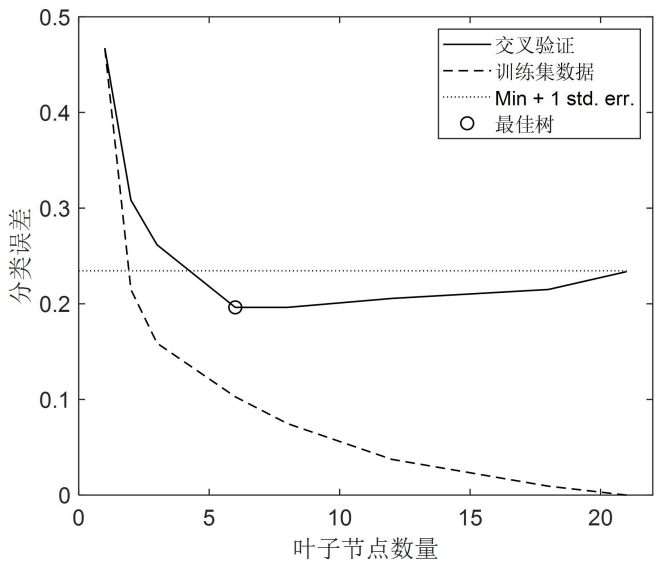

图 5 剪枝树规模（叶子节点数量）与分类误差的关系曲线

Fig.5 Relationship between the size of pruned tree (number of leaf nodes) and the misclassification error rate

决策树通过 MATLAB 2019b 建立，所得的一系列剪枝树的规模与分类误差关系如图 5 所示。可以发现，剪枝树对训练集数据的分类误差率随决策树规模减少而单调递增，表明饱和树对训练集数据的拟合效果是最好的。另一方面，剪枝树的交叉验证分类误差随决策树规模减少先减小后迅速增加。当叶子节点数量为 6 时，相应的剪枝树是距离最小分类误差一个标准差范围内规模最小的树，因此被选为最佳决策树。

最佳决策树如图 6 所示，它有 11 个节点，6 个叶子节点，深度为 3，交叉验证分类误差为 0.1963。类似文献[18,19]的最佳决策树分类误差落在 0.2 至 0.24 之间，说明本文的最佳决策树的分类准确度达到要求。

图 6 中每个长方形代表一个节点，呈现了节点中各项分值所含样本量与所占比例。在不同节点连接路径上的指标为分裂器名称，下方的数字表示父节点分裂为子节点后，Gini 系数的下降值。

最佳决策树的根节点通过分裂器购票与充值服务分裂为节点 1 和节点 2，分裂的指标得分阈值为 4.5 分。划分后，节点 1 和节点 2 的 Gini 系数之和与根节点相比降低了 0.210。节点 1 表示购票与充值服务得分均值小于等于 4.5 分的数据集，节点 2 则表示购票与充值服务得分均值大于 4.5 分的数据集。节点 1 和 2 继续沿其他分裂器分裂，直至叶子节点 3、4，7、8、9 和 10。

## 3.2 DT-IPA 决策结果

利用实证研究所获数据建立的 IPA 如图 7 所示，按 1.2.3 所述方法可以确定各个评价指标的初始改进优先级。由图 7 可知，评价指标车厢拥挤度、站内拥挤度和购票与充值服务位于优先改进区，具有最高的改进优先级。根据 1.2.4 所述方法求得这三个指标的改进可执行度分别为 0.06，0.19 和 0.74，可以发现车厢拥挤度改进可执行度较差。从现实情况分析，购票与充值服务可以通过增设机器数量从而减少乘客购票与充值的等待时间提升，站内拥挤度的服务质量可以通过限流进站提升，但由于高峰运营时段列车的发车频率已接近饱和状态且列车的容量一定，车厢拥挤度难以通过增大发车频率等运营管理手段降低，增购车辆的成本则可能过高，因此改进车厢拥挤度服务质量的策略可执行度较差，于是把车厢拥挤度记作改进可执行度不足指标。

表 2 呈现了从最佳决策树中提取的 6 条分类规则，它们均满足 S、P 和 Po 的最小阈值，是有效的。已知车厢拥挤度是改进可执行度不足的指标，根据 1.2.5 所述方法，序号 6 的分类规则可以用于调整由 IPA 确定的指标初始改进优先级并量化被调整指标的改进幅度。

6 号分类规则包含的指标为购票与充值服务、生命与财产安全和车厢拥挤度。由图 7 可知，生命与财产安全得分均值等于 4.5 分，购票与充值服务得分均值低于 4.5 分，两者均未满足 6 号分类规则所述要求，服务质量需要提升。根据 1.2.5 所述方法，购票与充值服务和生命与财产安全的改进优先级调整为第一级别，车厢拥挤度的改进优先级则调整为第四级别。

购票与充值服务得分均值差 0.29 分到达 4.5 分，而生命与财产安全得分均值若提高 0.01 分即可大于 4.5 分，从而满足 6 号分类规则。基于 6 号分类规则可以发现，即使不提升车厢拥挤度的服务质量，通过至少提高购票与充值服务 0.29 分和生命与财产安全 0.01 分，总体感知质量将有 94.3%的概率达到满分。经 DT-IPA 模型决策的长沙地铁服务质量提升策略总结如表 3。由表 3 可知，属于 DT-IPA 模型第一改进优先级的评价指标的改进可执行度高于 IPA 的决策结果，反映了 DT-IPA 模型的有效性。

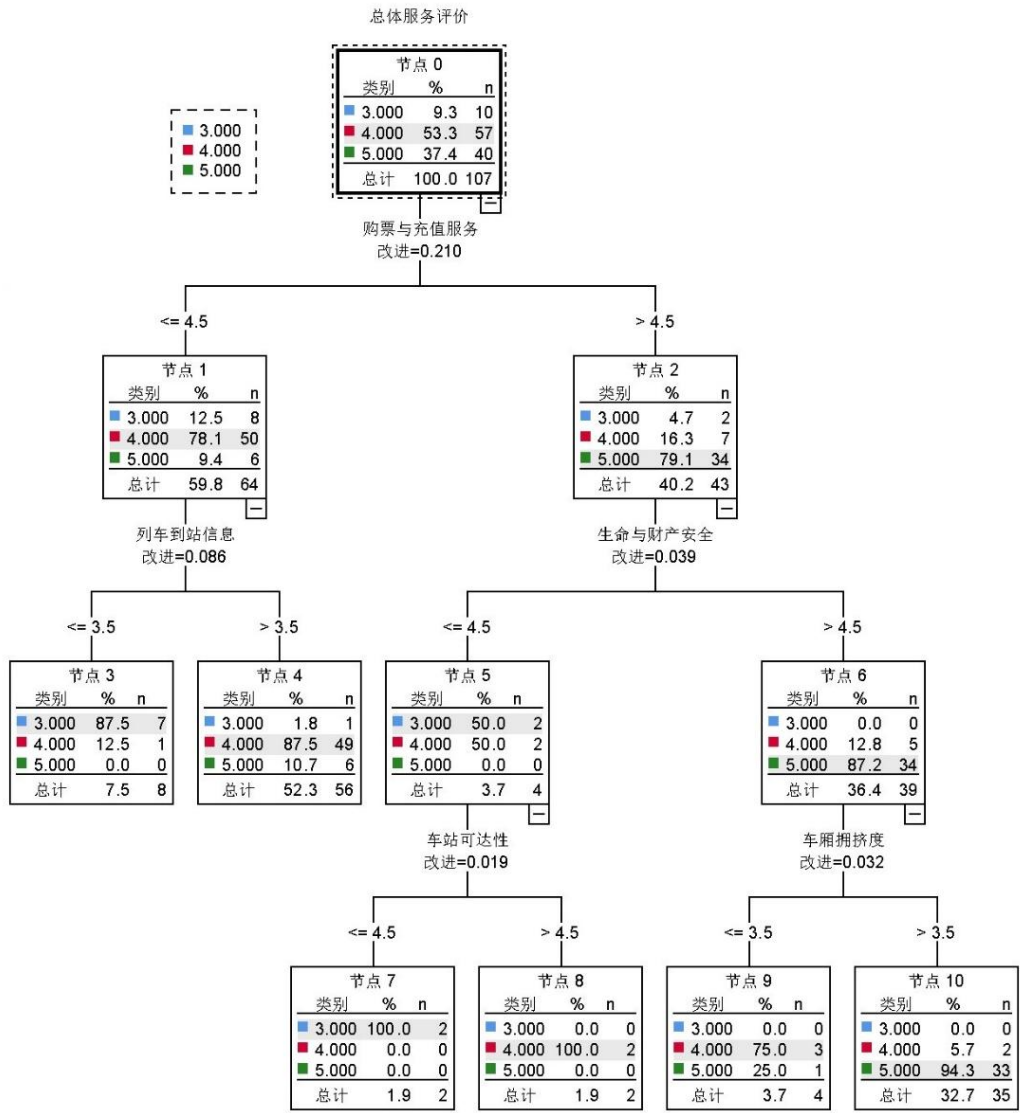

图 6 最佳决策树

Fig.5 The optimal decision tree

表 2 最佳决策树包含的分类规则

Table 2 If-then rules from the optimal decision tree

| 序号 | 如果（If） | 则（Then） | S | Po | P | 备注 |
|---|---|---|---|---|---|---|
| 1 | 购票与充值服务（≤4.5）且 列车到站信息（≤3.5） | 总体服务质量（=3） | 0.065 | 0.075 | 0.875 | 有效 |
| 2 | 购票与充值服务（≤4.5）且 列车到站信息（>3.5） | 总体服务质量（=4） | 0.458 | 0.523 | 0.875 | 有效 |

| 3 | 购票与充值服务（＞4.5）且 生命与财产安全（≤4.5）且 车站可达性（≤4.5） | 总体服务质量（=3） | 0.019 | 0.019 | 1 | 有效 |
| 4 | 购票与充值服务（＞4.5）且 生命与财产安全（≤4.5）且 车站可达性（＞4.5） | 总体服务质量（=4） | 0.019 | 0.019 | 1 | 有效 |
| 5 | 购票与充值服务（＞4.5）且 生命与财产安全（＞4.5）且 车厢拥挤度（≤3.5） | 总体服务质量（=4） | 0.028 | 0.037 | 0.750 | 有效 |
| **6** | **购票与充值服务（＞4.5）且 生命与财产安全（＞4.5）且 车厢拥挤度（＞3.5）** | 总体服务质量（=5） | 0.308 | 0.327 | 0.943 | 有效 |

注：S= Support，P= Probability，Po= Population，计算结果保留至千分位

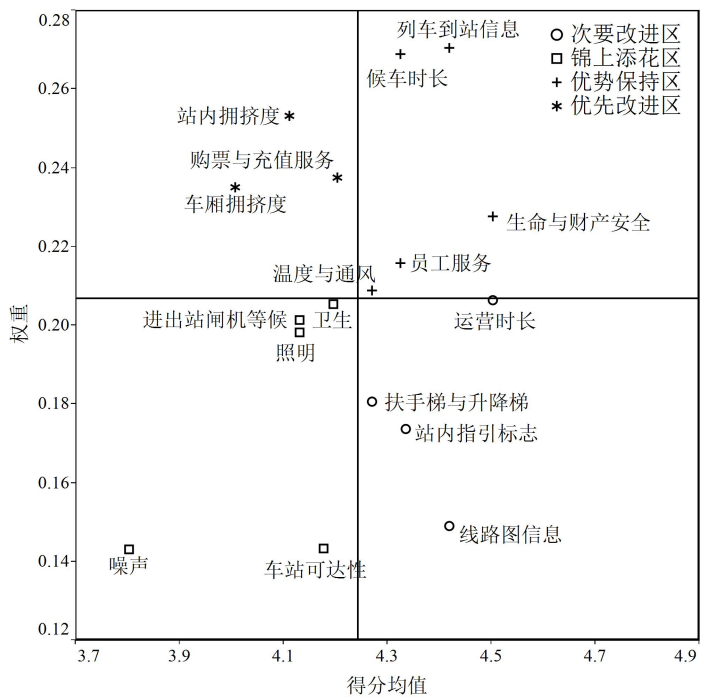

图 7 IPA 决策结果

Fig.7 The result of IPA

表 3 DT-IPA 模型决策结果

Table 3 The result of DT-IPA model

| 指标 | 改进优先级 | | 改进幅度 |
| --- | --- | --- | --- |
| | IPA | DT-IPA | |
| 车站可达性 | 第三 | 第三 | - |
| 站内指引标志 | 第四 | 第四 | - |
| 购票与充值服务 | 第一 | **第一** | ≥0.29分 |
| 进出站闸机等候 | 第三 | 第三 | - |
| 线路图信息 | 第四 | 第四 | - |
| 扶手梯与升降梯 | 第四 | 第四 | - |
| 站内拥挤度 | 第一 | 第一 | - |
| 列车到站信息 | 第二 | 第二 | - |
| 候车时长 | 第二 | 第二 | - |
| 车厢拥挤度 | 第一 | **第四** | - |
| 噪声 | 第三 | 第三 | - |
| 照明 | 第三 | 第三 | - |
| 温度与通风 | 第二 | 第二 | - |
| 卫生 | 第三 | 第三 | - |
| 员工服务 | 第二 | 第二 | - |
| 生命与财产安全 | 第二 | **第一** | ≥0.01分 |
| 运营时长 | 第四 | 第四 | - |

## 4 结论

1）与传统的 IPA 决策模型相比，本文构建的 DT-IPA 决策模型，融合考量了地铁服务质量指标的得分、权重及改进可执行度，使服务质量提升策略在一定时期内具有更好的可执行性。

2）运用层次分析法量化指标改进的可执行度时，主要通过专家采用标度法确定指标的相对可执行度，简单易行，但具有一定的主观性。

3）进一步的研究可以考虑采用相对客观的方法确定评价指标改进可执行度，进一步完善 DT-IPA 模型。

# Research on Metro Service Quality Improvement Schemes Considering Feasibility

*CHEN Weiya*[1]　　*LI Jiajia*[1]　　*KANG Zixuan*[1]

（1. School of Traffic and Transportation Engineering, Rail Data Research and Application Key Laboratory of Hunan Province, Central South University, Changsha 410075, China）

**Abstract:** It is an important management task of metro agencies to formulate reasonable improvement schemes based on the result of service quality surveys. Considering scores, weights, and improvement feasibility of service quality attributes in a certain period, this paper integrates Decision Tree (DT) into Importance-Performance analysis (IPA) to build a DT-IPA model, which is used to determine the improvement priority of attributes, and to quantify the improvement degree. If-then rules extracted from the optimal decision tree and the improvement feasibility computed by analytic hierarchy process are two main items derived from the DT-IPA model. They are used to optimize the initial improvement priority of attributes determined by IPA and to quantify the degree of improvement of the adjusted attributes. Then, the overall service quality can reach a high score, realizing the operation goal. The effectiveness of the DT-IPA model was verified through an empirical study which was taken place in Changsha Metro, China. The proposed method can be a decision-making tool for metro agency managers to improve the quality of metro service.

**Key words:** metro; service quality; improvement strategies; feasibility; decision tree